# 3D Iterative Spatiotemporal Filtering for Classification of Multitemporal Satellite Data Sets

Hessah Albanwan, Rongjun Qin, Xiaohu Lu, Mao Li, Desheng Liu, and Jean-Michel Guldmann

**Abstract**

*The current practice in land cover/land use change analysis relies heavily on the individually classified maps of the multitemporal data set. Due to varying acquisition conditions (e.g., illumination, sensors, seasonal differences), the classification maps yielded are often inconsistent through time for robust statistical analysis. 3D geometric features have been shown to be stable for assessing differences across the temporal data set. Therefore, in this article we investigate the use of a multitemporal orthophoto and digital surface model derived from satellite data for spatiotemporal classification. Our approach consists of two major steps: generating per-class probability distribution maps using the random-forest classifier with limited training samples, and making spatiotemporal inferences using an iterative 3D spatiotemporal filter operating on per-class probability maps. Our experimental results demonstrate that the proposed methods can consistently improve the individual classification results by 2%–6% and thus can be an important postclassification refinement approach.*

## Introduction

In land cover/land use change mapping, obtaining consistent data and classification results is critical for remote sensing applications and time-series analysis. Varying acquisition conditions—such as meteorological conditions, viewing angles, sun illumination, and sensor characteristics—largely affect the radiometric consistency through different temporal data sets. State-of-the-art satellite imaging platforms, including the Ikonos (1999, decommissioned), GeoEye-1 (2008), and WorldView satellite constellations, are able to collect very-high-resolution (VHR) multispectral images, with a spatial resolution that reaches as high as 0.3 m, which dramatically increases the capability to interpret and monitor land surfaces and urban/natural dynamics at a high level of detail. For instance, VHR imagery is much less affected by mixed-pixel effects than are low-resolution images, allowing information extraction to be carried out on an individual-object basis, such as shapes, sizes, and patterns of buildings, trees, roads, and so on. This certainly brings advantages for manual interpretation, but unfortunately the algorithmic development of automatic interpretation has not yet achieved equivalent advancement. As a result, the large volume of data to be robustly interpreted brings a data crisis for operators and decision-makers when enjoying the high resolution. The major technical challenge in classification/detection algorithms on VHR data is to overcome the low intraclass (within-class) similarity and interclass (between-classes) separability (Salehi, Zhang and Zhang 2011). As the spectral ambiguities of VHR images increase, the scene contents become more complex (Qin and Fang 2014)—for instance, as the images are clear enough to reflect individual objects such as buildings, pavements, ground, roads, and so on (Kotthaus et al. 2014), it is possible that these object classes may reflect similar spectral responses, since they are mainly made of similar materials (e.g., concretes). Therefore, classification algorithms considering spectral information alone might likely yield misclassifications (Lu, Hetrick and Moran 2010). Shape information of the objects in addition to the spectral responses may bring more discriminative powers, and relevant methods have achieved considerable improvement (Ghamisi, Dalla Mura and Benediktsson 2015; Qin 2015). However, such methods often suffer from unreliable segmentation and inconsistent spectral information, and they can be particularly vulnerable when applied to multitemporal data sets, where the varying spectral responses affect both segmentation and classification.

Existing studies have shown that integrating the spectral and 3D geometric information (e.g., height, depth) reduces the uncertainty of classification and change detection for objects with similar spectral characteristics (Chaabouni-Chouayakh et al. 2010; MacFaden et al. 2012; Qin, Tian and Reinartz 2016). Unlike image information, where appearances vary over time due to illumination and acquisition conditions, height information is relatively robust, providing more valuable information for multitemporal data comparison. With the growing number of optical satellite sensors in operation, the possibilities for satellites to view an area from multiple angles have dramatically increased. This presents digital surface models (DSMs) generated from such multiview or stereo-view data as common sources for mapping and monitoring. Compared to a typical lidar-based DSM, satellite-derived DSMs are more affordable and can capture images of inaccessible regions by aerial vehicles, but they present a higher level of noise. Thus, to better utilize such DSM data, the uncertainties of the data and the derived results (i.e., classification) need to be well accounted for.

In this article, we introduce a 3D iterative spatiotemporal filter that applies probability inference to refine land cover classification results of VHR multitemporal data. The proposed iterative spatiotemporal filter further extends past work (Albanwan and Qin 2018), which showed that a single-step

Hessah Albanwan and Xiaohu Lu are with the Geospatial Data Analytics Laboratory, Department of Civil, Environmental and Geodetic Engineering, The Ohio State University, Columbus, OH 43210.

Mao Li is with the Department of Electrical and Computer Engineering, The Ohio State University, Columbus, OH 43210.

Rongjun Qin is with the Geospatial Data Analytics Laboratory, Department of Civil, Environmental and Geodetic Engineering, and the Department of Electrical and Computer Engineering, The Ohio State University, Columbus, OH 43210 (qin.324@osu.edu).

Desheng Liu is with the Department of Geography, The Ohio State University, Columbus, OH 43210.

Jean-Michel Guldmann is with the Knowlton School of Architecture, The Ohio State University, Columbus, OH 43210.

spatiotemporal filter has the capability to improve the spectrum consistency across a multitemporal data set. The proposed method further enhances such capability for multimodal time-series data (i.e., spatial, spectrum, and height) for classification problems, which may achieve higher classification accuracy through inference through a multitemporal data set rather than independently classified images. Our proposed method is applied to per-class probability maps and can adaptively reduce the heterogeneity of the probability maps, leading to more consistent classification results through time. The rest of this work is organized as follows: The next section briefly introduces relevant works and the rationale of our proposed work. After that, we present the proposed 3D iterative spatiotemporal filter. The experimental results and evaluations are shown in the following section. The final section concludes by discussing the effectiveness and limitations of our work, and potential future improvements.

## Related Works and Rationale

The proposed work aims to enhance classification accuracy through inferring information across multitemporal data sets, and thus is very relevant to both classification problems and multitemporal data processing in a land use/land cover change (LULCC) context. Therefore, this section briefly summarizes both of these two relevant topics.

### Related Works

*Classification of VHR Satellite Images*
VHR image classification has been intensively investigated in remote sensing (Fauvel et al. 2013). Traditional classification methods are categorized into pixel- or object-based methods. In most cases, object-based methods perform reasonably better on VHR images due to the availability of the spatial information that allows use of patterns and local regions, which are often hard to obtain through pixel-based approaches (Weih and Riggan 2010; Chen et al. 2011; Keyport et al. 2018). The major challenge in classifying multitemporal VHR data sets is obtaining consistent classification maps through time, which can be difficult due to varying factors such as season changes, atmosphere, and acquisition conditions. Object-based methods often require segmentation of the image to extract spatial information (Qin et al. 2015). However, improper segmentation parameters can lead to under- or oversegmentation, which affects classification accuracy. Incorporating 3D geometric information (i.e., height) can provide more accurate, robust, and stable solutions, since the height of objects is insensitive to spectral variations (Minh and Hien 2011; MacFaden et al. 2012; Qin et al. 2016). Including height in the classification involves fusing multisource data (i.e., spectral and height information; Huang, Zhang and Gong 2011; Salehi et al. 2011; Kim 2016), which can be performed at either the pixel, feature, or decision level (J. Zhang 2010). One of the challenges in including height in classification is its source and quality. For instance, DSMs are either generated from direct measurements from sensors like lidar or computed using modern photogrammetric techniques (e.g., stereo matching). Although in general lidar data are more accurate, DSMs derived from images are preferred in many studies due to their higher availability and lower cost (Salehi et al. 2011). Stereo-matching algorithms used to generate DSMs depend on the quality of the pair of images and their acquisition conditions and thus may result in certain noises and uncertainties (Minh and Hien 2011). Current approaches mostly adapt statistical assumptions to the noises and either introduce new sources of data by fusing information from multispectral images or perform empirical statistical filtering (e.g., morphological processing; Chaabouni-Chouayakh et al. 2010; Moser, Serpico and Benediktsson 2013; Qin et al. 2015), which is capable of reducing certain noises although is still problematic when the content of the scene is complex (Fu 2011; Yan, Shaker and El-Ashmawy 2015).

*Spatiotemporal Inferences of Land Use/Land Cover Change Maps*
Land use/land cover change products are highly dependent on the quality of land cover classification maps. Inconsistencies and uncertainties in the classification maps due to the varying data spectra can negatively influence applications that directly rely on them, such as time-series analysis, change detection, object recognition, and modeling. In such a context, improving single-date classification may, on one hand, require ad hoc techniques specifically tailored for factors influencing the images and might, on the other hand, not yield optimally consistent classification maps through time. In this regard, spatiotemporal inference approaches applied on the images or classification results can be particularly effective, as they simultaneously homogenize information in both spatial and temporal directions (Floberg and Holden 2013). Current spatiotemporal inference models are categorized into local, nonlocal, and global models.

Local and nonlocal methods are more efficient in terms of computational complexity, since they operate over small pixel neighborhoods and local regions (Floberg and Holden 2013). For instance, Cheng et al. (2017) have proposed a spatial and temporal nonlocal filter-based fusion model to predict the land cover class for every pixel based on the spectral values through a weighted sum of the neighboring pixels in the data set. Albanwan and Qin (2018) proposed a local spatiotemporal bilateral filter as a preprocessing step for multitemporal images to enhance their radiometric characteristics and consistency for classification. Global approaches for spatiotemporal inference are mainly based on probabilistic graphical models and Markov random fields, which model each pixel as a statistical variable where local smoothness and global consistency serve as objectives when performing maximal likelihood estimation (Kasetkasem and Varshney 2002; Liu and Cai 2012; Gu, Lv and Hao 2017). Such inference algorithms help to generate more consistent change maps and prevent noise caused by inaccurate classification results. The solution for such inference algorithms usually uses fixed spatial-weight parameters, which might potentially lead to oversmoothing for the change maps. Gu et al. (2017) have proposed a linear weighting scheme, where the spatial weights are estimated adaptively for every pixel based on its changed, unchanged, or uncertain-change status. Although global methods used in change detection (such as Markov random fields) provide high-quality results, they involve per-pixel processing, which increases the computational complexity exponentially as the number of pixels and time-series data increase.

### Proposed Method and Rationale
The goal of this work is to achieve an efficient spatiotemporal inference model that allows accurate land use/land cover mapping. Inspired by work by Albanwan and Qin (2018), where a 3D spatiotemporal filter was developed for classification of images using spectral, spatial, and temporal knowledge, we propose here a simple but effective extension leading to a 3D iterative spatiotemporal filter. Using global methods for per-pixel classification is a time-consuming process. Krähenbühl and Koltun (2011) suggest using iterative approaches through local filters to approximate global inferences, with the benefit of reducing computational costs. Therefore, we apply here an iterative approach that filters the probability maps along with the corresponding DSMs in an iterative fashion to achieve optimal classification results. Figure 1 illustrates our rationale of using spatiotemporal filtering: Given three per-class probability maps generated from three temporal data sets, the highest probability normally determines the class of the pixel (represented as $c_i$ for the ith class). In the first and third images, the yellow pixel is classified as the second class $c_2$ and the second image is classified as $c_n$. Knowing that, then if the feature of the central pixel in the second image is dissimilar to that of the first and the third, this central pixel in the second image can be a misclassification. Probability inferences in the temporal direction might help correct such errors by providing higher confidence in the correct class. Our proposed iterative solution is expected to perform such probability inferences in a global sense, to refine and improve the consistency of the classification maps and their accuracy.

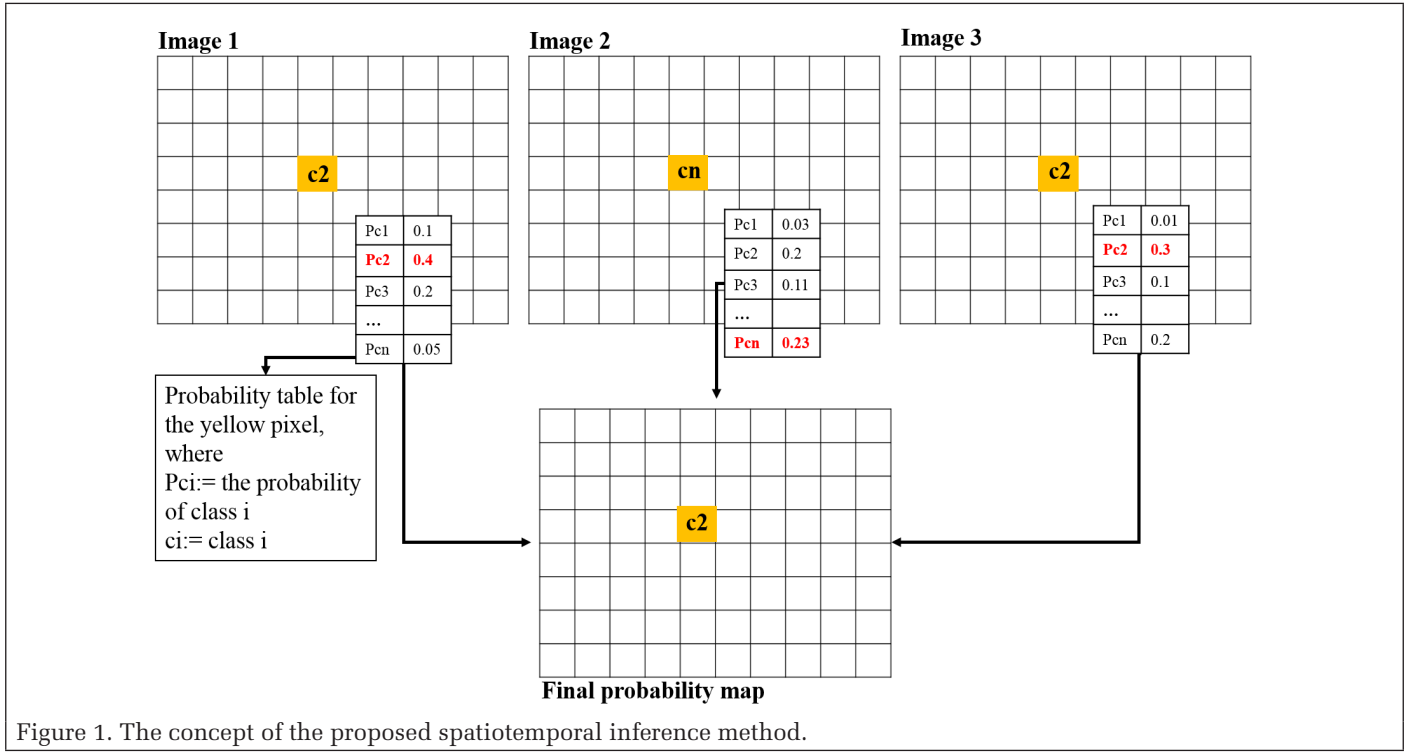

Figure 1. The concept of the proposed spatiotemporal inference method.

## Methodology

Our workflow consists of three main steps (Figure 2): data preprocessing of multitemporal stereo satellite images, object-based classification, and 3D iterative filtering for spatiotemporal inference. In this section, we provide an overview of the first two steps and emphasize details of the third, wherein our main technical contribution lies. The preprocessing step mainly consists of DSM and orthophoto generation and registration using stereo satellite images. Once the DSM and orthophotos are registered, we take advantage of their differences to infer training labels from one temporal data set to others, and then utilize these labels to train random-forest classifiers individually to generate initial and per-class probability maps. These probability maps are then fed into our proposed 3D iterative spatiotemporal filter, where probability maps are globally inferred (concept as per described in Figure 1) as the final probability map. We assess the classification results by estimating the overall accuracy of each iteration.

### Data Preprocessing

The preprocessing stage involves the generation of the DSM and orthophoto data sets, precise geometric alignment of the data, and normalization to nDSM. We generate high-quality DSMs and orthophotos of the related satellite image pairs using RPC Stereo Processor (Qin 2016), which adopts a hierarchical semiglobal matching method (Hirschmüller 2005). Hierarchical semiglobal matching offers a great benefit in being computational and time efficient; it reduces memory consumption by searching only a limited number of disparity ranges. Due to systematic satellite positioning error, the generated DSMs and orthophotos may not be well aligned. Co-registration of a DSM requires estimating the transformation parameters (rotation and translation) between the reference and target DSMs, as described by Gruen and Akca (2005). Since the rotation differences for a satellite overview data set are often regarded as negligible (Waser et al. 2008), we can then compute only the shift parameters in three directions x, y, and z. Using a simple approximation to least-squares surface matching (Gruen and Akca 2005), we estimate the shift parameters that minimize the sum of squared Euclidean distances. We eliminate the systematic error by discarding the potential noise and outliers due to the DSM generation process by applying a threshold (in our experiment we used 6 m with an empirical estimation of the height uncertainties), which regards points with an error greater than the threshold as outliers and discards them for error computation. Finally, we generate nDSM using a morphological top-hat reconstruction strategy as explained by Qin and Fang (2014).

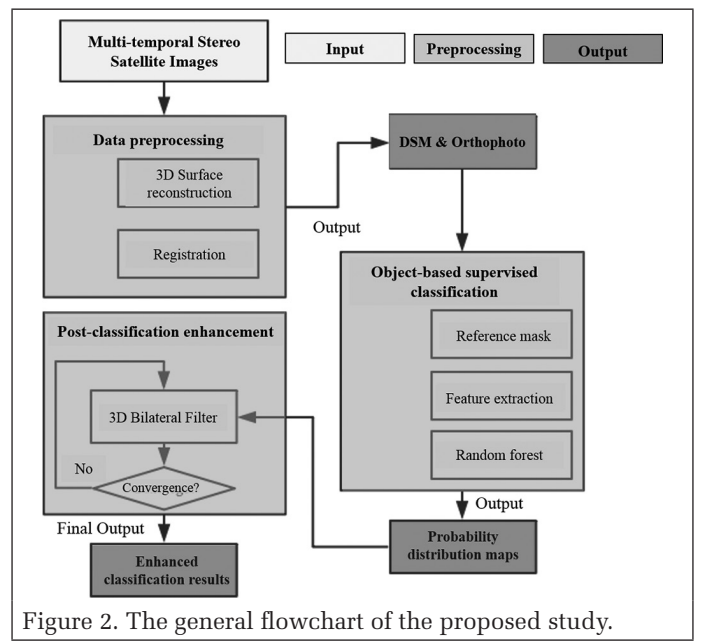

Figure 2. The general flowchart of the proposed study.

### Object-Based Classification and Initial Probability-Map Generation

*Training Label Generation Through Empirical Inferences*
Since generating training labels for every image of different dates is tedious and time-consuming, we simplify this process by propagating training samples for the entire data set through an intuitive and automated training-label mask generation approach. This empirically sets thresholds for image/DSM differences to determine whether or not to keep certain training labels from a labeled image of one date through time: Given a labeled image (i.e., the reference map), our automated method utilizes preliminary features such as nDSM and the normalized difference vegetation index as change indicators and determines if a pixel in the labeled images should be kept in the other temporal data set. The nDSM offers higher robustness against spectral variations that result from different illumination and atmospheric disturbances between the multitemporal data. This is equivalent to an intuitive threshold-based change detection method, where differences in both nDSM and the normalized difference vegetation index are used as indicators. Theoretically, these training labels should be generated manually for each data set, to ensure that they are evenly distributed and well balanced. In our procedure, this is ad hoc and can vary with data sets. However, we do set conservative thresholds to ensure minimal label inference errors, and we are aware that such a labeling mechanism may lead to suboptimal training-label sets. We therefore have verified the training labels generated and have performed the necessary manual refinement.

*Feature Extraction, Classification, and Probability-Map Generation*
Per-class probability (sometimes called confidence) maps are important to determine the class membership of each data point. This is provided by most of the statistical classifiers: For example, support-vector machines weight the distances of the output to the support boundaries, and random forests compute the probability by counting the votes from all the decision trees forming the forest. In our experiment, we use the random-forest classifier (Breiman 2001) and adopt an object-based classification approach, where mean shift (Comaniciu 2002) is used as the segmentation approach due to its wide use in relevant applications. The spectral feature is computed using principal-component analysis on the spectrum bands. We have also considered spatial features extracted from the DSMs and orthophotos as part of the feature vector, which includes dual morphological profiles and morphological top-hat reconstruction (Qing Zhang 2006; Qin et al. 2015). Details of these features are shown in Table 1.

As discussed, the random-forest classifier is an ensemble classifier that decides the labels based on the number of votes per class among the decision trees, and the normalized voting numbers (rationing the total number of votes) are then used as the probability/chance that a segment or pixel belongs to a specific class; 500 decision trees were used in our experiment, and their per-class probability maps yield in total N (number of classes) probability maps for each temporal data set.

### The Proposed 3D Iterative Spatiotemporal Filter for Probability Inference

The proposed filter is based on a 3D spatiotemporal bilateral filter (Albanwan and Qin 2018), the general form of which is represented as

$$P_{\text{new}}(x_i, y_i, t_m) = \frac{1}{N*T} \sum_{j \in N, n \in T} W_{3D}(x_j, y_j, t_n) * P(x_j, y_j, t_n), \quad (1)$$

where N denotes the spatial neighborhood centered on point $(x_i, y_i)$ and T refers to the number of observations (dates) through time; $P(x, y, t_n)$ is the probability of the pixel at location $(x, y)$ in the probability map taken on date $t_n$; and $W_{3D}$ are adaptive weights that consider spectral, spatial, and temporal differences in the data sets. Since in our context the DSM data are available and regarded as robust to illumination changes, we hereby take the nDSM differences as means of weighting the temporal coherences for probability inference. Thus, the weight $W_{3D}$ can be decomposed into three components:

$$W_{3D} = W_{\text{spatial}} * W_{\text{spectral}} * W_{\text{nDSM}}, \quad (2)$$

where

$$W_{\text{spatial}}(x_j, y_j, t_n) = e^{-\frac{\|x_i - x_j\|^2 + \|y_i - y_j\|^2}{2\sigma_s^2}} \quad (3)$$

$$W_{\text{spectral}}(x_j, y_j) = e^{-\left(\frac{\|I(x_i, y_i) - I(x_j, y_j)\|^2}{2\sigma_r^2}\right)}. \quad (4)$$

The $\sigma_s$, $\sigma_r$, and $\sigma_h$ parameters are the spatial, spectral, and elevation/temporal bandwidths of the weighted filter; $I(x_i, y_i)$ refers to the values of the color image (transformed color space); and nDSM refers to the normalized height values. In the spatial and spectral domains, the pixels are weighted based on their closeness and spectral similarities to the central pixel in a defined window, where larger weights are assigned to spectrally similar and spatially proximate pixels (and vice versa). The value of $\sigma_s$ is empirically determined based on the size of the window; a large value can lead to an overly smoothed image with blurry edges. The parameter $\sigma_r$ serves as the spectral bandwidth of a typical bilateral filter to allow edge-aware filtering (Tomasi and Manduchi 1998). CIELAB color is used as the transformed color space for computing the spectral differences, where three bands (near-infrared, red, and green) are used for the color transformation (Joblove and Greenberg 1978; Tomasi and Manduchi 1998). The height variation is class-dependent: For example, the ground class normally has smaller variations, thus it should have a smaller bandwidth to penalize larger weight for incorporating changed areas in the filtering processing. Also, tree classes normally have a higher uncertainty, and thus a larger bandwidth is needed to average out noises. Therefore, the relevant bandwidth $\sigma_h$ here is estimated per class to achieve optimal performance. The height range of each class is approximated by the labeled pixels in each of the training labels; for a certain class c, the elevation bandwidth $\sigma_h$ and the relevant weight are computed using the following equations:

$$\sigma_h = \left(\frac{70\%}{2}\right) * [\text{value range of nDSM of class } c] \quad (5)$$

$$W_{\text{nDSM}}(x_j, y_j, t_n) = e^{-\frac{\|\text{nDSM}(x_i, y_i, t_m) - \text{nDSM}(x_j, y_j, t_n)\|^2}{2\sigma_h^2}} \quad (6)$$

Table 1. Extracted features used for classification, the light boxes are types of features, and the dark boxes under the same columns provide additional details.

| Principal-Component Analysis (PCA) | Dual Morphological Profiles (DMP) and Morphological Top-Hat Reconstruction of Digital Surface Model | Adaptive Structural Elements |
|---|---|---|
| All columns from PCA of the spectrum (which includes only the RGB bands) of segment S (Wold et al. 1987; Jolliffe 2005). | The radius sequence of disk shape used in morphological profile reconstruction [24 146 260] pixels. DMP of the first component from PCA in the first feature. DMP for the color inverse with the first component from PCA in the first feature. | Used to construct DMP and address the problem of multiscale information of segments (Qian Zhang et al. 2015). |

The 70% value comes from the consideration that the height measurements from stereo-matching introduce noise. We therefore take the one-sigma rule to determine $\sigma_h$ and, by assuming the height-difference variation as a zero-centered Gaussian distribution, compute the one-sigma interval as 68.27% ≈ 70% of the nDSM. These all together give the final form of the 3D weight as

$$W_{3D}(x_j, y_j, t_n) = e^{-\left(\frac{\|x_i - x_j\|^2 + \|y_i - y_j\|^2}{2\sigma_s^2} + \frac{\|I(x_i, y_i) - I(x_j, y_j)\|^2}{2\sigma_r^2} + \frac{\|\text{nDSM}(x_i, y_i, t_m) - \text{nDSM}(x_j, y_j, t_n)\|^2}{2\sigma_h^2}\right)}. \quad (7)$$

This 3D weighted filter performs local spatiotemporal filtering in a limited receptive field for each pixel. To achieve optimal probability maps, we further globalize the inference algorithm through iterative application to probability maps. The iterative filtering is global by means of gradually propagating information from locally processed cells to wider receptive fields. We use the probability maps generated from the random-forest classifier (as described in the previous subsection) as the initial probability maps (see Figure 4) for inference. For each date, there will be in total n (number of classes) probability maps, and the spatiotemporal inference is performed on probabilities of the same classes. In the filtering process, the total weight $W_{3D}$ for each pixel is constant through all iterations, since it is a function of the spectral, spatial, and height information of the image. The probability maps are updated in each iteration:

$$P_c^k(x_i, y_i, t_n) = \frac{1}{N*T} \sum \sum_{j \in N, n \in T} W_{3D}(x_j, y_j, t_n) * P_c^{k-1}(x_j, y_j, t_n) \quad (8)$$

where $P_c^k(x_i, y_i, t_n)$ is the estimated probability of pixel $(x_i, y_i)$ at date $t_n$, in class c, on the kth iteration, derived from the probability $P_c^{k-1}(x_j, y_j, t_n)$ on iteration $(k-1)$. We compute this iterative process until the difference between $P_c^{k-1}$ and $P_c^k$ is smaller than a threshold, which in this work is set as 5% of relative changes:

$$\text{convergence criterion} = \frac{P_c^k(x_i, y_i, t_n) - P_c^{k-1}(x_i, y_i, t_n)}{P_c^k(x_i, y_i, t_n)} * 100\% < 5\%. \quad (9)$$

We take the processed probability map and determine the class label for every pixel as the one with the highest probability. Using the ground-truth data, we can evaluate the accuracy of the resulting classification maps to understand how the proposed iterative spatiotemporal filter improves the results. The computational complexity for every iteration is $O(W H T)$, where W and H are the width and height of the probability map and T is the number of temporal data. Figure 3 presents the pseudocode of the algorithm.

## Experimental Results

### Data Description

The data set used in our experiment is a multitemporal data set in Port-au Prince, Haiti, through the 2010 earthquake, when a catastrophic earthquake with magnitude 7.0 $M_w$ caused a large number of fatalities and extreme damages to the area, forcing Haitians to migrate into temporary (tents) and long-term lodging (i.e., shelters last for a longer time for accommodations). Therefore, our classification work particularly includes changes in buildings with different functionalities (i.e., long-term lodgings, temporary lodgings, and normal built-up areas). The satellite data set contains seven on-track stereo pairs (data collected on the same day with the satellite on the same track) from 2007 to 2014 and one incidental image pair in 2015 over Port-au-Prince, Haiti, with data-acquisition date and details shown in Table 2.

We selected three test regions with a size of 1 × 1 km² (i.e., 2001 × 2001 pixels for images resampled at 0.5 ground-sampling distance) in various urban scenes (see Figure 4 for details). Test region 1 is an open area around the airport in which refugees and the government built up their temporary and long-term lodgings spontaneously. We consider seven classes in the classification: buildings, ground, trees, roads, grass, temporary lodging, and long-term lodging (see Figure 4). The three test regions vary in terms of their urban forms: Test region 1 indicates an open area surrounded by dense buildings, test region 2 is specifically dense residential areas with narrow streets, and test region 3 is a school zone with moderate and larger buildings among dense and small buildings. The selection of the three test regions is based on our visual inspection for representative regions across the data set, and apparent surface changes were observed in these regions as well, which will be quantified in detail in subsequent sections.

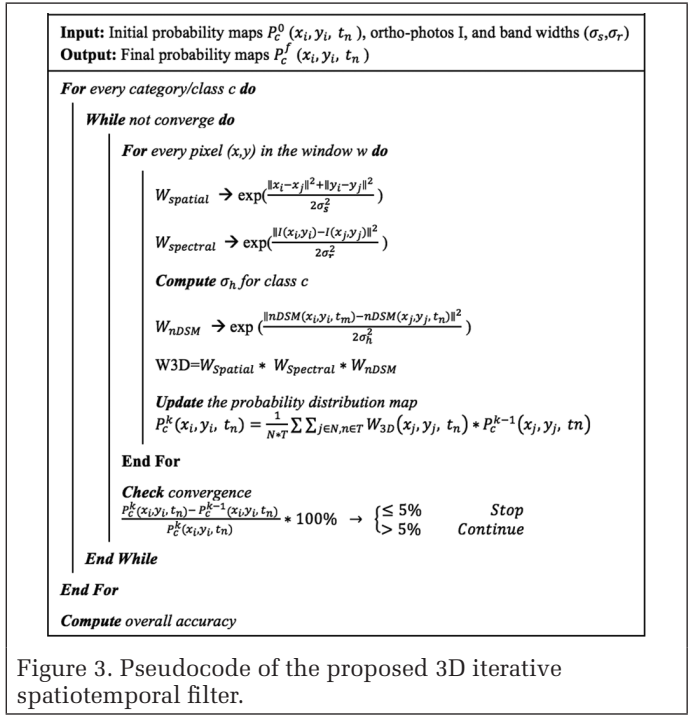

Figure 3. Pseudocode of the proposed 3D iterative spatiotemporal filter.

Table 2. List of satellite stereo images used in this work.

| Satellite | Date | Resolution (Panchromatic; m) | Resolution (Multispectral; m) |
|---|---|---|---|
| *Ikonos* | 3/22/2007 (on track) | 0.82 | 3.28 |
| *GeoEye-1* | 1/16/2010 (on track) | 0.41 | 1.84 |
| *Ikonos* | 6/6/2010 (on track) | 0.82 | 3.28 |
| *Ikonos* | 12/21/2010 (on track) | 0.82 | 3.28 |
| *GeoEye-1* | 3/8/2012 (on track) | 0.41 | 1.84 |
| *GeoEye-1* | 9/11/2013 (on track) | 0.41 | 1.84 |
| *WorldView* | 7/24/2014 (on track) | 0.46 | 1.84 |
| *GeoEye-1* | 6/23/2015, 7/1/2015 | 0.41 | 1.84 |

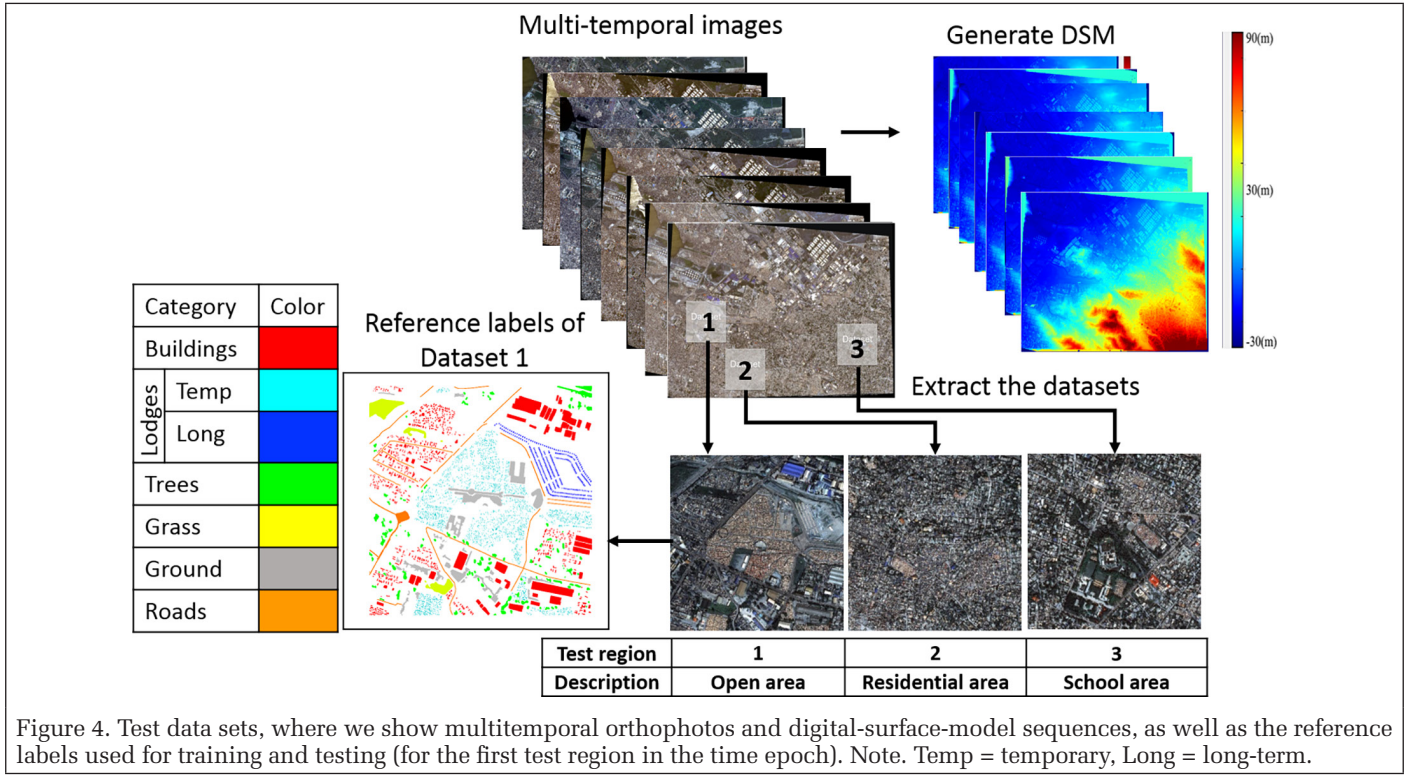

Figure 4. Test data sets, where we show multitemporal orthophotos and digital-surface-model sequences, as well as the reference labels used for training and testing (for the first test region in the time epoch). Note. Temp = temporary, Long = long-term.

**Data Preprocessing:** DSM **and Orthophoto Generation**
In this preprocessing step, we generate the orthophoto and DSMs for the entire area of Port-au-Prince, Haiti, using the RPC Stereo Processor software and the stereo satellite imagery. The DSM generation is based on the semiglobal matching algorithm (Hirschmüller 2005). The the DSMs are geometrically aligned and nDSMs are generated, as described earlier. Figure 4 shows an overview of these data sets. During this step, all data sets with varying spatial resolutions are sampled to the highest cell size through the data sets (0.5 m).

**Classification and Probability-Map Generation**
The initial object-based classification is generated using methods as described in previous sections. We keep the per-class probability map (already described) for further spatiotemporal inference. This initial classification yields a relatively satisfactory classification map to start with, with the overall accuracy shown in Table 3. In general, test region 2 (dense residential regions) yields the lowest overall accuracy (around 80%), primarily due to the complex rooftops and potentially higher uncertainties of the DSM.

**Experimental Details of the Proposed 3D Iterative Spatiotemporal Filter**
*Parameters*
As already described3, the 3D iterative spatiotemporal filter has spatial, spectral, and temporal components (height difference through the temporal data set). Based on the described rationale of determining these bandwidth parameters, we apply the following parameters based on our data set.

Table 3. Overall initial classification accuracy for the three test regions.

| | Accuracy (%) | | |
|---|---|---|---|
| Date | Test region 1 | Test region 2 | Test region 3 |
| 3/Mar 2007 | 91.04 | 83.47 | 91.12 |
| 1/Jan 2010 | 93.21 | 81.50 | 93.06 |
| 6/Jun 2010 | 91.93 | 83.52 | 88.82 |
| 12/Dec 2010 | 89.08 | 80.81 | 88.58 |
| 3/Mar 2012 | 92.19 | 81.43 | 91.44 |
| 9/Sep 2013 | 90.40 | 81.03 | 94.99 |
| 7/Jul 2014 | 95.11 | 82.19 | 90.39 |
| 7/Jul 2015 | 92.74 | 83.22 | 94.61 |

**Bandwidth for the spectral and spatial domains**: Because of the size of each test image (over 2000 × 2000 pixels), the window size is set to 5 × 5 pixels to balance the efficiency and accuracy of the results. The bandwidth $\sigma_s$ for each dimension of the spatial domain in the filter window is set to 3. In addition, for 8-bit images (ranging from 0 to 255) the bandwidth of the spectral domain $\sigma_r$ is set empirically to 5, and this is shown in the original bilateral filter (Tomasi and Manduchi 1998) to have good leverage on smoothing and edge preserving.

**Bandwidth for temporal (height) range domain** $\sigma_h$: The elevation bandwidth is determined separately for each class as specified in Equation 6. Table 4 shows the bandwidth $\sigma_h$ for each class, calculated using statistics in the test regions; this is to be done for each test region, as they have their own training labels. We can see that $\sigma_h$ for buildings and trees is much higher than for the ground and grass classes. This is reasonable because the building and tree classes, in general, have larger height variations, and for the same amount of height difference the ground class should have smaller bandwidth to be sensitive to large changes, while the trees should have moderate tolerance to height differences when considering weighted summing probability maps.

Table 4. Temporal (height) domain bandwidth $\sigma_h$ for each class on test region 1.

| Item | Height (m) |
|---|---|
| Buildings | 5.98 |
| Temporary lodging | 0.64 |
| Long-term lodging | 1.50 |
| Ground | 0.53 |
| Trees | 4.45 |
| Grass | 0.56 |
| Roads | 1.08 |

*Visual and Statistical Analysis of the Proposed Iterative Spatiotemporal Filter*
**Visual analysis**: Our method enhances classification by utilizing the per-class probabilities of time-series data via spatiotemporal inference. This is reasonable, as unchanged objects can be used to mutually enhance the fidelity of the classification probability maps. We compare the initial and enhanced classification results for all the data sets in Figure 5. For instance, the mislabeled objects in the initial classification map of the dense areas in test regions 2 and 3 are corrected after being processed by the proposed spatiotemporal inference method. As shown in regions within the black circle, open grounds primarily dominated by temporary lodgings are misclassified to the ground or grass classes in the initial classification map, and the proposed spatiotemporal inference method shows that these misclassifications can be effectively corrected and demonstrates that classification maps are visually consistent with the images (quantifiable accuracy improvement shown under Statistical analysis).

To demonstrate how the iterative process improves the classification results, we select a small patch from the date of 2014/07 in test region 1 (Figure 6) to analyze their classification map in each iteration of the process. The patch shows two buildings with close adjacency, and its initial classification map (Figure 6b) shows the enlarged view where the buildings are partly misclassified to long-term lodgings. As can be seen from the figures, the iterative process gradually recovers the classification map where, in the 11th iteration (convergence), the final result (Figure 6g) shows a much more complete building segment.

**Statistical analysis**: Table 5 shows the classification accuracies of each image in each test region before and after their probability maps are processed by the proposed method, where "Before" refers to the overall accuracy of initial classification results, "After" indicates the enhanced accuracy through our proposed method, and "Δ" represents the accuracy change in the enhanced results over the initial ones. The average increase in the accuracy of the three test regions is 4.2363%, 5.285%, and 4.723%, respectively.

In Table 6 we took a small patch of pixels in test region 2 to show probability changes with the number of iterations. The first patch is a part of a building area, and the second patch shows trees. In both patches, we notice that the probability of the correct class increases with iterations (see the arrows) and those of the remaining classes decrease.

Figure 7 presents the overall accuracy for the entire set of iterations and three experiments. We can see in Figure 7 that the first iteration plays a major role in the accuracy enhancement. In all three experiments, the first iteration increased about 4%–6% of the overall accuracy and the rest of the iterations (until convergence) contribute approximately 1%–2%. On average, we note that all experiments converge around the fifth iteration.

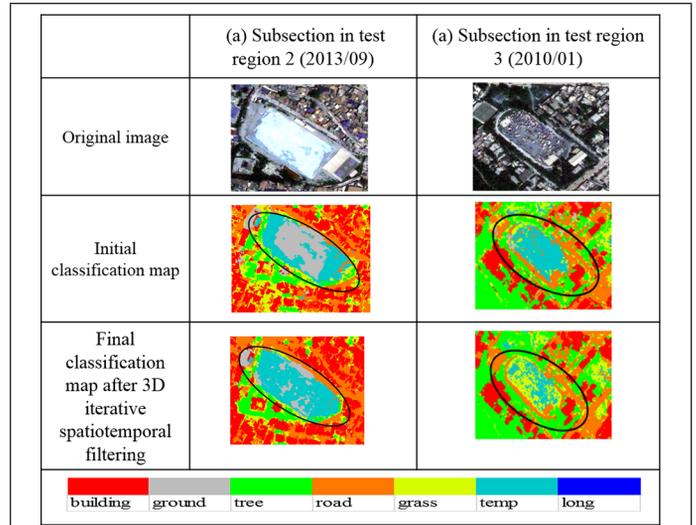

Figure 5. Two sample patches in test regions 2 and 3 illustrating the enhancement of classification after the proposed filter.

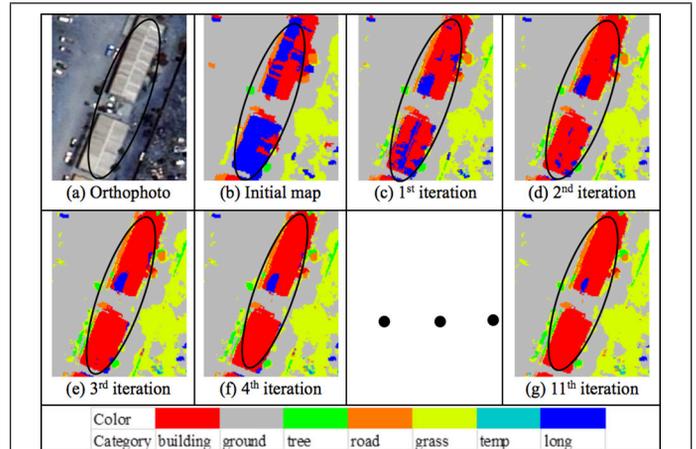

Figure 6. Detailed classification result for each iteration of enhancement for a patch in test region 1.

Table 5. The overall accuracy (%) before and after probability enhancement for each test region. (Note:. the bold numbers indicate indicates the maximum increase in the accuracy.)

| | Test region 1 | | | Test region 2 | | | Test region 3 | | |
|---|---|---|---|---|---|---|---|---|---|
| Date | Before | After | Δ | Before | After | Δ | Before | After | Δ |
| 2007 | 91.04 | 95.21 | +4.17 | 83.47 | 88.14 | +4.67 | 91.12 | 95.85 | +4.73 |
| 1/Jan 2010 | 93.21 | 96.45 | +3.24 | 81.50 | 85.67 | +4.17 | 93.06 | 96.82 | +3.76 |
| 6/Jun 2010 | 91.93 | 96.26 | +4.33 | 83.52 | 89.79 | **+6.27** | 88.82 | 94.87 | **+6.05** |
| 12/Dec 2010 | 89.08 | 95.57 | **+6.49** | 80.81 | 87.59 | **+6.78** | 88.58 | 94.86 | **+6.28** |
| 3/Mar 2012 | 92.19 | 95.92 | +3.73 | 81.43 | 86.92 | +5.49 | 91.44 | 97.08 | +5.64 |
| 9/Sep 2013 | 90.40 | 96.56 | **+6.16** | 81.03 | 87.29 | **+6.26** | 94.99 | 97.54 | +2.55 |
| 7/Jul 2014 | 95.11 | 97.27 | +2.16 | 82.19 | 88.90 | +6.17 | 90.39 | 96.58 | +6.19 |
| 2015 | 92.74 | 96.35 | +3.61 | 83.22 | 85.69 | +2.47 | 94.61 | 97.19 | +2.58 |
| Average | 92.09 | 96.20 | +4.24 | 82.15 | 87.50 | +5.29 | 91.63 | 96.35 | +4.72 |

## Conclusion and Future Work

We have proposed a spatiotemporal inference method that improves the classification accuracy of multiview and multitemporal VHR satellite images. Knowing that temporal classification maps often suffer from inconsistencies in their results due to highly variable features that change over time, we propose using 3D geometric features (i.e., DSMs generated from multiview images) for better stability. We applied our proposed spatiotemporal inference method to process VHR imagery and DSMs to enhance the classification accuracy of each individual temporal data set. Our proposed method approximates the global inference model through iterative processing of 3D bilateral filters utilizing the spatial, spectral, and geometric information. Our method achieved a 2% to 6% increase in overall accuracy for all experiments with varying scene complexities. Notable improvements were observed in certain scenarios. For example, for the open (sparse) areas in the first and third data sets, the accuracy increased from ≈89% to ≈97%, whereas for dense regions in test region 2, the accuracy increased from ≈80% to ≈90%. Since the proposed method is a postprocessing method operating on probability maps, it is able to work on probability maps generated by different types of classifiers as long as the probability maps can be derived. In addition, the proposed method can be further generalized to less-restrictive data sets by considering the temporal component to be the image themselves. For instance, it can be applied on moderate- and low-resolution satellite imagery and can be used to process video sequences or different types of time-series raster data to enhance the image data qualities.

Table 6. Mean probabilities (%) for each class changed through iterations using a sample patch of buildings and trees in test region 2.

| Iteration | Class | | | | | |
| --- | --- | --- | --- | --- | --- | --- |
| | Buildings | Grass | Ground | Road | Temp | Tree |
| **Buildings** | | | | | | |
| 1 | 73.16 | 1.68 | 17.69 | 11.53 | 0.92 | 0.24 |
| 2 | **74.95↑** | 1.85 | 16.36 | 10.65 | 1.09 | 0.25 |
| 3 | **75.45↑** | 1.96 | 15.87 | 10.28 | 1.19 | 0.25 |
| 4 | **75.53↑** | 2.05 | 15.70 | 10.11 | 1.25 | 0.26 |
| 5 | 75.53 | 2.14 | 15.70 | 10.11 | 1.30 | 0.27 |
| 6 | 75.53 | 2.14 | 15.70 | 10.11 | 1.35 | 0.27 |
| 7 | 75.53 | 2.14 | 15.70 | 10.11 | 1.35 | 0.28 |
| **Trees** | | | | | | |
| 1 | 0.12 | 24.66 | 0.23 | 0.01 | 0.01 | 80.69 |
| 2 | 0.10 | 23.76 | 0.22 | 0.01 | 0.01 | **82.97↑** |
| 3 | 0.09 | 23.18 | 0.22 | 0.01 | 0.01 | **84.03↑** |
| 4 | 0.08 | 22.75 | 0.22 | 0.01 | 0.01 | **84.58↑** |
| 5 | 0.08 | 22.41 | 0.22 | 0.01 | 0.01 | **84.86↑** |
| 6 | 0.08 | 22.41 | 0.22 | 0.01 | 0.01 | **84.95↑** |
| 7 | 0.08 | 22.41 | 0.22 | 0.01 | 0.01 | 84.91 |

## Acknowledgments
This work is supported by the Sustainable Institute at The Ohio State University. The authors would like to thank the anonymous reviewers for their comments and suggestions that led to the final version of the manuscript.

## References
Albanwan, H. and R. Qin. 2018. A novel spectrum enhancement technique for multi-temporal, multi-spectral data using spatial-temporal filtering. *ISPRS Journal of Photogrammetry and Remote Sensing* 142:51–63.

Chaabouni-Chouayakh, H., T. Krauss, P. d'Angelo and P. Reinartz. 2010. 3D change detection inside urban areas using different digital surface models. *The International Archives of the Photogrammetry, Remote Sensing and Spatial Information Sciences* 38 (3B):86–91.

Chen, G., G. J. Hay, L.M.T. Carvalho and M. A. Wulder. 2012. Object-based change detection. *International Journal of Remote Sensing* 33 (14):4434–4457.

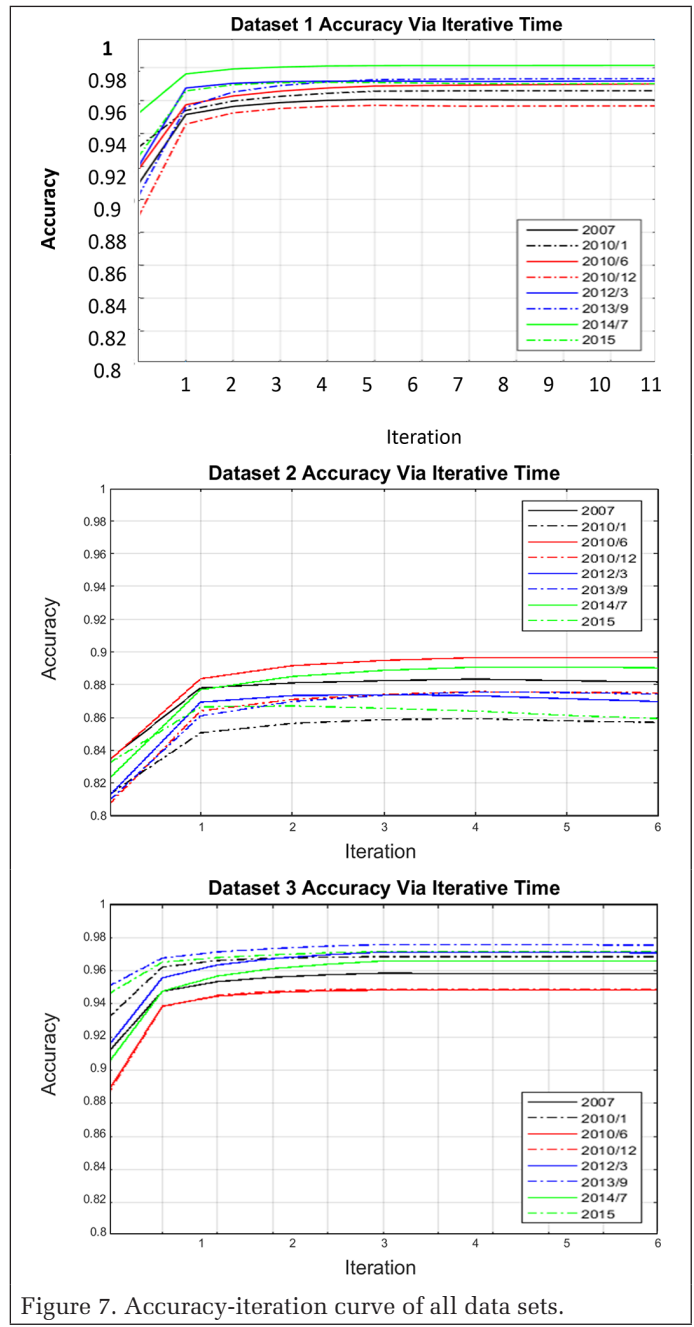

Figure 7. Accuracy-iteration curve of all data sets.

Cheng, Q., H. Liu, H. Shen, P. Wu and L. Zhang. 2017. Spatial and temporal nonlocal filter-based data fusion method. *IEEE Transactions on Geoscience and Remote Sensing* 55:4476–4488.

Fauvel, M., Y. Tarabalka, J. A. Benediktsson, J. Chanussot and J. C. Tilton. (2013). Advances in spectral-spatial classification of hyperspectral images. *Proceedings of the IEEE* 101 (3):652–675.

Floberg, J. M. and J. E. Holden. 2013. Nonlinear spatio-temporal filtering of dynamic PET data using a four-dimensional Gaussian filter and expectation-maximization deconvolution. *Physics in Medicine and Biology* 58 (4):1151–1168.

Fu, T.-C. 2011. A review on time series data mining. *Engineering Applications of Artificial Intelligence* 24 (1):164–181.

Ghamisi, P., M. Dalla Mura and J. A. Benediktsson. 2015. A survey on spectral–spatial classification techniques based on attribute profiles. *IEEE Transactions on Geoscience and Remote Sensing* 53 (5):2335–2353.

Gruen, A. and D. Akca. 2005. Least squares 3D surface and curve matching. *ISPRS Journal of Photogrammetry and Remote Sensing* 59 (3):151–174.

Gu, W., Z. Lv and M. Hao. 2017. Change detection method for remote sensing images based on an improved Markov random field. *Multimedia Tools and Applications* 76 (17):17719–17734.

Hirschmüller, H. 2005. Accurate and efficient stereo processing by semi-global matching and mutual information. Pages 807–814 in *Proceedings, 2005 IEEE Computer Society Conference on Computer Vision and Pattern Recognition*, Vol. 2, held in San Diego, California, 20–25 June 2005. Edited by C. Schmid, S. Soatto and C. Tomasi. Los Alamitos, California: IEEE Computer Society.

Huang, X., L. Zhang and W. Gong. 2011. Information fusion of aerial images and LIDAR data in urban areas: Vector-stacking, re-classification and post-processing approaches. *International Journal of Remote Sensing* 32 (1):69–84.

Joblove, G. H. and D. Greenberg. 1978. Color spaces for computer graphics. *ACM SIGGRAPH Computer Graphics* 12 (3):20–25.

Jolliffe, I. 2005. Principal component analysis. In *Encyclopedia of Statistics in Behavioral Science*, Wiley Online Library.

Kasetkasem, T. and P. K. Varshney. 2002. An image change detection algorithm based on Markov random field models. *IEEE Transactions on Geoscience and Remote Sensing* 40 (8):1815–1823.

Keyport, R. N., T. Oommen, T. R. Martha, K. S. Sajinkumar and J. S. Gierke. 2018. A comparative analysis of pixel- and object-based detection of landslides from very high-resolution images. *International Journal of Applied Earth Observation and Geoinformation* 64:1–11.

Kim, Y. 2016. Generation of land cover maps through the fusion of aerial images and airborne LiDAR data in urban areas. *Remote Sensing* 8 (6):521.

Kotthaus, S., T.E.L. Smith, M. J. Wooster and C.S.B. Grimmond. 2014. Derivation of an urban materials spectral library through emittance and reflectance spectroscopy. *ISPRS Journal of Photogrammetry and Remote Sensing* 94:194–212.

Krähenbühl, P. and V. Koltun. 2011. Efficient inference in fully connected CRFs with Gaussian edge potentials. Pages 109–117 in *Proceedings of the 24th International Conference on Neural Information Processing Systems*, held in Granada, Spain, 12–15 December 2011. Edited by J. Shawe-Taylor, R. S. Zemel, P. L. Bartlett, F. Pereira and K. Q. Weinberger.

Liu, D. and S. Cai. 2012. A spatial-temporal modeling approach to reconstructing land-cover change trajectories from multi-temporal satellite imagery. *Annals of the Association of American Geographers* 102 (6):1329–1347.

Lu, D., S. Hetrick and E. Moran. 2010. Land cover classification in a complex urban-rural landscape with QuickBird imagery. *Photogrammetric Engineering and Remote Sensing* 76 (10):1159–1168.

MacFaden, S. W., J.P.M. O'Neil-Dunne, A. R. Royar, J.W.T. Lu and A. G. Rundle. 2012. High-resolution tree canopy mapping for New York City using LIDAR and object-based image analysis. *Journal of Applied Remote Sensing* 6 (1):63567.

Minh, N. Q. and L. P. Hien. 2011. Land cover classification using LiDAR intensity data and neural network. *Journal of the Korean Society of Surveying, Geodesy, Photogrammetry and Cartography* 29 (4):429–438.

Moser, G., S. B. Serpico and J. A. Benediktsson. 2013. Land-cover mapping by Markov modeling of spatial–contextual information in very-high-resolution remote sensing images. *Proceedings of the IEEE* 101 (3):631–651.

Qin, R. 2015. A mean shift vector-based shape feature for classification of high spatial resolution remotely sensed imagery. *IEEE Journal of Selected Topics in Applied Earth Observations and Remote Sensing* 8 (5):1974–1985.

Qin, R. 2016. RPC Stereo Processor (RSP)—A software package for digital surface model and orthophoto generation from satellite stereo imagery. *ISPRS Annals of the Photogrammetry, Remote Sensing and Spatial Information Sciences* III-1:77–82.

Qin, R. and W. Fang. 2014. A hierarchical building detection method for very high resolution remotely sensed images combined with DSM using graph cut optimization. *Photogrammetric Engineering and Remote Sensing* 80 (9):873–883.

Qin, R., X. Huang, A. Gruen and G. Schmitt. 2015. Object-based 3-D building change detection on multitemporal stereo images. *IEEE Journal of Selected Topics in Applied Earth Observations and Remote Sensing* 8 (5):2125–2137.

Qin, R., J. Tian and P. Reinartz. 2016. Spatiotemporal inferences for use in building detection using series of very-high-resolution space-borne stereo images. *International Journal of Remote Sensing* 37 (15):3455–3476.

Salehi, B., Y. Zhang and M. Zhong, M. 2011. Object-based land cover classification of urban areas using VHR imagery and photogrammetrically-derived DSM. Pages 461–467 in *American Society for Photogrammetry and Remote Sensing Annual Conference 2011*, held in Milwaukee, Wisconsin, 1–5 May 2011. Edited by J. Editor. Bethesda, Maryland: American Society for Photogrammetry and Remote Sensing.

Tomasi, C. and R. Manduchi. 1998. Bilateral filtering for gray and color images. Pages 839–846 in *Sixth International Conference on Computer Vision*, held in Bombay, India, 4–7 January 1998. Edited by J. Editor. New Delhi, India: Narosa Publishing House.

Waser, L. T., E. Baltsavias, K. Ecker, H. Eisenbeiss, E. Feldmeyer-Christe, C. Ginzler, M. Küchler and L. Zhang. 2008. Assessing changes of forest area and shrub encroachment in a mire ecosystem using digital surface models and CIR aerial images. *Remote Sensing of Environment* 112 (5):1956–1968.

Weih Jr, R. C. and Riggan Jr, N. D. 2010. Object-based classification vs. pixel-based classification: Comparative importance of multi-resolution imagery. T*he International Archives of the Photogrammetry, Remote Sensing and Spatial Information Sciences* XXXVIII-4/C7:PPP–PPP.

Yan, W. Y., A. Shaker and N. El-Ashmawy. 2015. Urban land cover classification using airborne LiDAR data: A review. *Remote Sensing of Environment* 158:295–310.

Zhang, J. 2010. Multi-source remote sensing data fusion: Status and trends. *International Journal of Image and Data Fusion* 1 (1):5–24.

Zhang, Q. [Qian], R. Qin, X. Huang, Y. Fang and L. Liu. 2015. Classification of ultra-high resolution orthophotos combined with DSM using a dual morphological top hat profile. *Remote Sensing* 7 (12):16422–16440.

Zhang, Q. [Qing]. 2006. An empirical evaluation of the random forests classifier models for variable selection in a large-scale lung cancer case-control study (Doctoral dissertation). Retrieved from ProQuest Dissertations and Theses database. (3259518)